\documentclass{article}

\usepackage{amsmath}
\usepackage{amssymb}

\usepackage{graphicx}

\usepackage{cite}

\usepackage{color} 

\usepackage[labelfont=bf,labelsep=period,justification=raggedright]{caption}

\makeatletter
\renewcommand{\@biblabel}[1]{\quad#1.}
\makeatother

\date{}

\begin{document}

% Title must be 150 characters or less
\begin{flushleft}
{\Large
\textbf{Ensemble Risk Modeling Method for Robust Learning on Scarce Data}
}
% Insert Author names, affiliations and corresponding author email.
\\
Marina Sapir, 
\\
 MetaPattern, Bar Harbor, ME
\\
$\ast$ E-mail: m@sapir.us
\end{flushleft}

% Please keep the abstract between 250 and 300 words
\section*{Abstract}

In  medical risk modeling,  typical data are ``scarce'': they have  relatively small  number of training instances (N), censoring,  and high dimensionality (M). We show that the problem may be effectively simplified by reducing it to bipartite ranking, and introduce new bipartite ranking algorithm, Smooth Rank, for robust learning on scarce data. The algorithm  is based on ensemble learning with  unsupervised aggregation of predictors.  The advantage of our approach is confirmed in comparison with two ``gold standard'' risk modeling methods on 10 real life survival analysis datasets, where the new approach has the best results on all but two datasets with the largest ratio N/M. For systematic study of the effects of data scarcity on modeling by all three methods, we conducted two types of computational experiments:  on real life data with randomly drawn training sets of different sizes, and on artificial data  with increasing number of features.  Both experiments demonstrated that Smooth Rank  has critical  advantage over the popular  methods on the scarce data; it does not suffer from overfitting where other methods do.

\section{Introduction}
\subsection{The Survival Analysis Problem}
Survival analysis  deals with   datasets, where each observation $d=\langle x, t, \delta \rangle$ includes   covariate vector $x \in \chi$, a  {\it survival time} $t \in \mathbb{R}, t > 0$ and a binary {\it event indicator} $\delta \in \{0, 1\}: \delta = 1$, if an event (failure) occurred, and $\delta= 0$ if the observation $d$  is  censored at time $t.$ So, the {\it survival time} means either time to event or to the end of study, if the observation was stopped ({\it censored} ) without the event.   The last two variables represent the {\it target} or outcome of the observation. 

For example, in a study about risk of metastases after cancer surgery, survival time is time from the surgery to the discovery of metastases, if they occurred, or to  some other end of study. The observations where metastases were not discovered will end up censored. The censoring may occur either because the cancer was removed by surgery, or because the patients lost to follow up before the cancer metastasized. Some of these lost patients could have died from different causes or moved to another location.   

The prediction in survival analysis is generally understood as an estimate of an individual's risk, and the concept of risk is associated not only with the fact of event, but with its timing: the earlier the event happened, the higher was the risk. 

Because of incomplete (censored) observations, the individuals in study can be  partially ordered by their risk: for observations $d_i = \langle x_i, t_i, \delta_i \rangle, d_j = \langle x_j, t_j, \delta_j \rangle, d_i \prec d_j \equiv \delta_i = 1 \, \& \, t_i < t_j.$ 

Denote $S: \chi \rightarrow\mathbb{R}$ a model built by a survival analysis algorithm. The commonly accepted  criterion  of the risk modeling is Harrell's concordance index (CI) \cite{Tutorial} between the  risk order $\prec$ and the values of  $S(x)$. The concordance index is based on the concept of  {\it concordant} pairs of observations:  such pairs  $\langle d_i, d_j \rangle$, where both $S(x_i) < S(x_j)$ and $d_i \prec d_j.$  The pairs, where $S(x_i) > S(x_j)$ and $d_i \prec d_j$ are called {\it discordant}. By definition,
$$CI = \frac{CP + 0.5 \cdot Ties}{CP + DP + Ties,} $$ where $CP, DP, Ties$ are numbers of concordant pairs, discordant pairs and ties, respectively. With continuos features, the ties are unlikely, and in absence of ties, concordance index measures proportion of concordant pairs. 

Thus, the goal can be formalized as learning the partially known order $\prec$  by the way of  building a scoring function $S(x)$ on $\chi$ to minimize the number of discordant pairs of observations.   This puts the survival analysis in the class of the  supervised ranking problems  (see, for example \cite{RankUstat, CortesRank}). 

What makes the problem difficult is the data quality in medical longitudinal studies, which we call ``scarcity'': the number of  observations tends to be small; many of them may be censored;  number of features may be large comparing with the number of observations, the outcome and quantitative features depend to a large degree on unknown  factors; clinical features are subjective; some of the features may be irrelevant  - and so on.  All this put main emphasis on the robustness of learning. 

\subsection{Traditional Approach}

The survival analysis is not commonly treated as a ranking problem. Usually,   the solution is found by  modeling the time it takes for events to occur {\cite{Segal, Survey2} (a more difficult problem), and   the most common method  for this purpose is Cox proportional hazard regression (Cox PH) \cite{Cox}. Cox PH regression builds a model by maximizing likelihood of the observed survival rates. Originally, the method was  intended to study dependence between few covariates and the outcome; if used for prediction, it  tends to overfit on typical data. Popular tutorial \cite{Tutorial} recommends that number of covariates for Cox regression does not exceed 1/10th of the number of uncensored observations. 

In machine learning, most of the developed approaches try to ``improve'' Cox PH regression one way or another. The most popular methods include $L_1$ or $L_2$ penalization  of the regression parameters \cite{L1penalized, L2penalized}, to make learning  more robust. As we show here on real and artificial data, this  regularization may not be sufficient, as it often does not lead to improved performance on scarce data and does not prevent overfitting. 

\subsection{Alternative Approach}
We propose an  alternative approach to the survival analysis learning. 

{\bf First}, we address the noise in outcome by defining the ``risk'' as a chance of having failure before certain time $\mathcal{T}$.  The interpretation of risk as a propensity of an individual to have an early failure is natural and acceptable for medical practitioners. Considering survival analysis as a ranking problem and splitting observations on two classes (``early failure'' vs ``no early failure'')  we reduce the problem to bipartite ranking (see, for example, \cite{Bipartit}).  Binarization of the survival outcome simplifies the problem and decreases influence of noise   in the survival times.  

Let us notice, that even though only the order between two classes is modeled, the model produces continuous scores associated with the levels of risk, which, in turn, is related with the timing of events. Therefore,   the  performance still can be evaluated by the concordance index between the scores and  the order   $\prec$ on the test data.   

{\bf Second}, we introduce new bipartite ranking method Smooth Rank, designed specifically for the scarce data. 
The method  is based on the {\em strong} regularization technique used in Naive Bayes:  unsupervised aggregation of independently built univariate predictors.  Avoiding multidimensional optimization makes Naive Bayes less sensitive to the ``curse of dimensionality'',  allows it to be competitive with more sophisticated methods  \cite{Friedman} on scarce data.   The term {\em strong} regularization was first used in \cite{Segal} with reference to weighted voting \cite{Golub}, which is based on the same approach.  

In addition to the strong regularization, Smooth Rank employs smoothing techniques to make the model more robust,  less dependent on peculiarities of the small training samples. 

\subsection{Comparison of the Two Approaches}
To show that our approach is working,  Smooth Rank is compared here with $L_1$-penalized path method CoxPath and the Cox PH regression on 10 real life survival analysis datasets.  Smooth Rank has the best performance on 8 datasets. It yields to other methods only on the two datasets with the largest $N/M$ ratio.
 
In addition, we study relationship between data scarcity and methods' performance in two types of computations experiments. First, on two real life datasets, we  randomly exclude some observations to produce series of training sets of different sizes. Then, on the artificial  datasets, we gradually increase the number of features, keeping the size of the training set constant. Both experiments demonstrate that Smooth Rank has significant advantage in performance on scarce data, while other methods may work better on ``rich'' data.

 %###########
 % here should be the plan of the article. The experiments are lost! 

% Results and Discussion can be combined.
\section{Smooth Rank}

\subsection{Definition of Smooth Rank}

The main scheme  of the algorithm can be described as two steps procedure:

\begin{enumerate}
\item Independently for each feature $x^i:$ build a predictor $f_i(x^i)$ and calculate its weight $w_i$ based on its  performance;
\item Calculate a scoring function $F(x) = \sum_1^n w_i \cdot f_i(x^i).$

\end{enumerate}

For  a classification problem, there are two popular ensemble algorithms which follow this scheme. One of them is Naive Bayes classifier \cite{HTF} where all weights     $w_i \equiv 1$ and  each predictor  is built as a log-ratio between densities  of two classes:  $f_i(x^i) = log(d^i_{1}(x^i) / d^i_{2}(x^i)).$  Another example of an algorithm with the same scheme is ``weighted voting''  \cite{Golub}. 

There are several ways the general scheme  can be implemented in the context of survival analysis. Below is one of the possible implementations.

The algorithm is applied to the data, where observations are split on two classes with labels $C \in \{1,2\}$ by  survival time threshold $\mathcal{T}$.\\

\textbf{Bipartite Ranking Algorithm Smooth Rank}
\begin{itemize}
\item For each feature $x^i$:
\begin{enumerate}
\item Build kernel approximations $g^i_{1}, g^i_2$ of the density  of each class on $R^i = dom(x^i)$;
\item For each point $r \in R^i$ calculate $$q_i(r) = \frac{ g^i_1(r)  -   g^i_{2}(r) }{\pi_1 \cdot g^i_1(r)  + \pi_2 \cdot g^i_{2}(r)}, $$ where $\pi_1, \pi_2$ are frequencies of the classes 1 and 2;
\item Build marginal predictors  $\widetilde{q_i}(x)$  as smooth approximation of the function $q_i(x)$
\item Calculate weights $w_i$ of the predictors based on their correlation with outcome
\end{enumerate}
\item Calculate scoring function
$$ F(x) = \sum_{i: \, x^i \neq NA} w_i \cdot \widetilde{q_i}(x^i) / \sum_{i: \, x^i \neq NA}  w_i $$
\end{itemize}

\subsection{Implementation and parameters}
The algorithm is implemented in R using some standard R functions. 

\subsubsection{Selection of the time threshold $\mathcal{T}$}
In the experiments, we select the time threshold $\mathcal{T}$ to make the classes similar in size. The class 1 contains the events only: $d_i=\{x_i, \delta_i, t_i\}: \delta_i = 1 \, \& \, t_i \le \mathcal{T}$. The class 2 contains all observations with the survival time above $\mathcal{T}.$ It means that the censored observations with censoring time below the threshold $\mathcal{T}$ are excluded from training. 

\subsubsection{Density evaluation}
The density is approximated with cosine kernel. The  R function \textit{density} \cite{Density} uses Fourier transform  with a discretized version of the kernel and then makes linear approximation to evaluate the density at the specified points. Density was evaluated with default function parameters, on  equally spaced 512 points.

%%%% add references %%%%%%%%%%%%%%%%%%%%%%%

\subsubsection{Building marginal predictors}

The function $q_i(r)$ is less sensitive to the errors in the density evaluation than the default function used in Naive Bayes: $g^i_1(r) /  g^i_{2}(r)$. However, for the areas where density of both classes is low, small errors in  $g^i_1, g^i_{2}(r)$ can lead to big errors in  $q_i(r)$.  To deal with this issue, $q_i(r)$ is not evaluated for  $r: \pi_1 \cdot g^i_1(r)  + \pi_2 \cdot g^i_{2}(r) < 0.1 $. The aggregation on the last step handles values of the predictors in these points as missing.
The function $q_i(r)$ is smoothed on the Step 3 using  \textit{loess} procedure with the default parameters.  LOESS stands for  ``locally weighted scatterplot smoothing'' \cite{LOESS}. Advantage of this method is that it does not require to specify the class of functions for approximation. The procedure \textit{loess} was used with polynomials of the  degree 1.

\subsubsection{Calculation of weights}

The calculation of weights is implemented as a two step procedure. 

First, weights of the predictors are calculated by the formula $$w_i = CI(\widetilde{q_i}(x)) - 0.5,$$  where $CI(y)$ is a concordance index between the variable $y$ and the outcome. For two-class outcome, in absence of ties, CI is equal area under the ROC curve, which is common performance measure for bipartite ranking \cite{CortesRank}. 

The next step includes ``post-filtering'', or ``shrinkage''. The goal of  this step is to improve learning on the datasets where  many correlated weak predictors can  overweight  few strong ones. Rather than setting a hard threshold for selection of predictors, or use data to optimize the threshold,  the filtering is made based on comparison of all weights with the highest weight $\mu = max( w_j)$. The updated weights are calculated by the formula:

$$ w_i := \left\{ \begin{array}{ll}
                   w_i - \mu / 3, & \,if \, \, w_i > \mu / 3 \\
                   0, & otherwise.
                   \end{array}
                   \right.
  $$
The empiric  formula allows to filter out relatively weak predictors, making the filtering data-dependent without  tuning of hyper-parameters. 

\subsection{Properties of the predictors}

Since $g^i_{1}, g^i_2$ approximate densities of both classes, according to Bayes theorem, the marginal conditional probability of the class $j$ in the point $x^i = r$ can be defined by the formula: $$ P(Y = j | x^i = r) \simeq \frac{\pi_j \cdot g^i_j(r) }{\pi_1 \cdot g^i_1(r)  + \pi_2 \cdot g^i_{2}(r)},$$ where $Y$ is the class labels, $\pi_1, \pi_2$ are priors for the two classes, approximated by their frequencies. Then the marginal predictor  $\widetilde{q_i}(r) $ can be presented as
$$\widetilde{q_i}(r) \simeq \frac{P(Y = 1 | x^i = r)}{\pi_1} - \frac{P(Y = 2 | x^i = r)}{\pi_2},$$ 
difference between ratios of conditional posterior probabilities to prior probabilities in two classes.  If variable $x^i$ is conditionally independent on  $Y$ in $x^i = r$, both  posterior probabilities  in this point are equal to their priors,   and $\widetilde{q_i}(r) \simeq 0.$  In each point $r$, the value of marginal predictor function $\widetilde{q_i}(r) $ indicates degree and direction of  local association between the values of the variable $x^i$ and the response variable.   Then, $\widetilde{q_i}(r) $ influences the scoring function $F$ only for those points $x^i = r$  which are predictive, and do not participate in ranking otherwise.  

The fact that densities for each class are evaluated independently increases robustness of the proposed function. 

In our implementation, the parameters of the kernel approximation and LOESS procedures were fixed to ensure maximal smoothness. Thus, unlike most of other advanced risk modeling methods, Smooth Rank does not have hyper-parameters to tune up. 

\section{Results on Real Data}

\subsection{Algorithms under comparison}
Smooth Rank is compared with two algorithms, which become standard in survival analysis \cite{Segal, Survey2, Segal08}.

\subsubsection{Cox proportional hazard regression}

In the traditional approach by sir David Cox \cite{Cox},   risk of failure  is understood as a time-dependent  ``hazard function''  $\Lambda(x, t)$:  cumulative probability of an individual $x$ having event (failure) up to time $t$. 

 Cox proportional hazard (PH) regression is based on the strong assumption that the hazard function has the form of $$\Lambda(x, t) = \lambda_0(t) \cdot  exp(\beta(x)),$$ where $\lambda_0(t)$ is  unknown time-dependent function, common for all individuals in the population. The assumption implies, in particularly, that for any two individuals, their hazards are proportional all the time.  
Accordingly, the result of the modeling is  not the time-dependent hazard functions, but rather the ``proportionality'' scores. 

The method can not be applied on data with $M > N. $

 We use the method's implementation from the R package \textit{survival}. The method does not work with missing values.

\subsubsection{CoxPath algorithm}

CoxPath \cite{Park} algorithm is one of most popular approaches to regularization of the Cox PH regression.  The path algorithm implements $L_1$-penalized Cox regression with series of values of regularization parameter $\lambda$. The important property of the $L_1$-regularization is that it includes automatic feature selection, and it can work when number of features exceeds number of training cases. 

The function implemented in  the R package \textit{glmpath} by the method's authors is used here. The function builds regression models at the values of $\lambda$ at which the set of non-zero coefficients changes.  For each model, the function outputs values of three criteria: AIC, BIC, loglik. The criterion AIC was chosen to select the best model for the given training set. We used default values of the parameters of the \textit{coxpath} procedure. The method does not work with missing values.

\subsection{The Datasets}

The next datasets were used for methods comparison. 

\begin{itemize}	
\item BMT:  The dataset represents data on 137 bone marrow transplant patients \cite{BMT} . The data allow to model several outcomes. Here,  the  models are built for disease free survival time. The first feature is diagnosis, which has three values:       ALL; AML Low Risk; AML High Risk. Other features characterize demographics of the patient and donor, hospital, time of waiting for transplant, and some characteristics of the treatment. There are 11 features overall, among them two are nominal. 

\item Colon: These are data from one of the first successful trials of adjuvant chemotherapy for colon cancer. Levamisole is a low-toxicity compound previously used to treat worm infestations in animals; 5-FU is a moderately toxic (as these things go) chemotherapy agent. There is possibility to model two outcome: recurrence and death. The data can be found in R package ÒsurvivalÓ.  The features include ÒtreatmentÓ (with three options: Observation, Levamisole, Levamisole+5-FU); properties of the tumor, number of lymph nodes. There are total 11 features and 929 observations. 

\item Lung1: Survival in patients with advanced lung cancer from the North Central Cancer Treatment Group \cite{Lung1}. Performance scores rate how well the patient can perform usual daily activities. Other features characterize calories  intake and weight loss. The dataset has 228 records with 7 features

\item  Lung2, the dataset from \cite{Lung2} Along with the patients' performance scores, the features include cell type (squamous, small cell, adeno, and large), type of treatment and prior treatment. 

\item BC : Breast cancer  dataset \cite{BC}. It contains 7 tumor characteristics in 97 records of patients. 

\item PBC : This data is from the Mayo Clinic trial in primary biliary cirrhosis of the liver conducted between 1974 and 1984 \cite{PBC}. Patients are characterized by standard description of the disease conditions.   The dataset has 17 features and 228 observations.

\item Al:  The data  \cite{Al} of the 40 patients with diffuse large
B-cell lymphoma contain information about 148 gene expressions associated with cell proliferation from ÒlympohichipÓ microarray data. Since there are more features than the observations, the Cox regression could not be applied on the data. 

\item Ro02s: the dataset from \cite{Ro02} contains information about 240 patients with lymphoma. Using hierarchical cluster analysis on whole dataset and expert knowledge about factors associated with disease progression, the authors identified relevant four clusters and a single gene out of the 7399  genes on the lymphochip. Along with gene expressions, the data include two features for histological grouping of the patients. The authors aggregated gene expressions in each selected cluster to create a ÒsignaturesÓ of the clusters. The signatures, rather than gene expressions themselves were used for modeling.  The dataset with aggregated data has 7 features. 

\item Ro03g, Ro03s: the data \cite{Ro03} of 92 lymphoma patients. The input variables include data from ÒlymphochipÓ as well as results of some other tests. The Ro03s data contain averaged values of the gene expressions related with cell proliferation (proliferation signature). The Ro03g dataset includes the values of the gene expressions included in the proliferation cluster, instead of their average. Thus, the Ro03s dataset contains 6 features, and the dataset Ro03g contains 26 features. 
\end{itemize}

\subsection{Description of the experiment}

For each dataset we did 100 random splits on train and test data in proportion $2:1$. For each split, all methods were applied on the train data and tested on the test data. Thus, the splits are the same for all the methods. The performance  of each method was evaluated by average concordance index  on the test data. 

Cox PH regression and CoxPath algorithms do not work with missing values, while Smooth Rank does. So, for the first two methods 
records with missing values were removed. The  exception is the data Al, where there are too few records and many missing values. In this dataset  5-nearest neighbors imputation was used for all the methods.

\subsection{The results}
\begin{table*}
\caption{Comparison of Methods on Survival Analysis Data}
\centering
\begin{tabular}{| l | l | l | l | l|  l |  l  | l |}
%\toprule
\hline
$\#$ & Data & N $\times$ M &  N/M & Smooth Rank & Cox & Cox Path   \\ 
%\midrule
\hline
1 & BMT  &  137 $\times$ 11 & 12.4  
	& \textbf{0.68} (6.4)
	& 0.58  
	& 0.58 
	
	 \\

2 & Colon & 929 $\times$ 11 & 84.4 
	& 0.65 (4)
	& \textbf{0.66}  
	&  \textbf{0.66} 
	
	 \\
	
3 & Lung1 &  228 $\times$ 7 & 32.6 
	& \textbf{0.63} (5.7)
	& 0.62  
	& 0.62  
	 \\ 

4 & Lung2 &  137 $\times$ 6 & 22.8 
	       & \textbf{0.73} (2.16)
		   & 0.69   
	       & 0.70  
	       \\

5 & BCW &  97 $\times$ 7 & 13.9 
	& \textbf{0.71}  (5.9)
	& 0.69 
	& 0.69  
	\\
	
6 & PBC & 418 $\times$ 17 & 24.6 
	& \textbf{0.83} (12.6)
	& 0.82  
	& 0.82 

	\\
	
7 & Al & 40 $\times$ 148 & 0.27 
	& \textbf{0.63} (110) 
	& ---  
	& 0.52 

	\\
		
8 & Ro02s  & 240 $\times$ 7 & 34.3 
	& 0.70 (7)
	& \textbf{0.73}  
	& \textbf{0.73}  
	 \\
	
9 & Ro03s & 92 $\times$ 6 & 15.3 
	& \textbf{0.76} (3)
	& 0.74  
	& 0.75  
	\\
	
10 & Ro03g &  92 $\times$ 26 & 3.54 
	& \textbf{0.76} (23)
	& 0.58 
	& 0.67 
	\\

%\bottomrule
\hline
\end{tabular}

{Every cell contains mean value of CI on the test data for 100 random splits in proportion 2:1.  The highest values in each row are marked by bold font. The number in brackets is average number of features left after filtering in Smooth Rank}

\end{table*}

The results are presented in the Table 1. The ratio $N/M$ is included as some measure of the dataset scarcity: the smaller is the ratio, the less representative (more scarce) is the dataset.  For all datasets, the table contains average CI for each method for each dataset.

The Table 1 shows that in 8 out of 10 cases Smooth Rank has the best results. Smooth Rank yields to two other methods on the 2 datasets (lines 2 and 8) with the largest ratio $N/M$. 

In the three cases with the lowest ratio of $N/M$ (lines 1,7,10) the advantage of Smooth Rank is the most prominent. Its performance is higher than performance of  other methods by $9\% - 11\%.$ 

Let us notice that the dataset R003s is a processed version of the dataset Ro03g: Ro03g contains original values of gene expression, and  Ro03s includes aggregated features, ``signatures''.   While Smooth Rank has equally good (the best) results with or without aggregation,  two other methods require preliminary feature aggregation for comparable performance.

The table allows to contrast two types of regularization: the traditional one, with $L1$- penalization, and the proposed here alternative approach, which includes the strong regularization. 

CoxPath uses penalization and model selection to improve Cox PH regression, but most of times it has almost identical accuracy with Cox PH regression on the given data. A possible explanation is that the optimal model selection on the same small training set as part of the path method leads to ``model selection bias'' \cite{ModelSelection, ModelSelectionBias}. 

Advantages in accuracy of Smooth Rank over other methods indicate superiority of the strong regularization for risk modeling on scarce data.

\section{Experiments with Controlled Data  Scarcity}

To confirm the advantages of Smooth Rank on scarce data, we conducted two series of experiments controlling two aspects of the data scarcity: number of training instances and dimensionality. In the first series with real data, some instances were randomly removed to obtain training sets of various sizes. In the second experiment on artificial data, the number of instances was fixed, but the number of features was gradually increased. The goal was to see the trends in methods performance on the series of modified datasets.   

\subsection{Experiments with reduced number of training cases}

The experiments were conducted on two of the largest real life datasets from our list,  PBC and Colon, where  missing values were imputed using 5 nearest neighbors. 

For each dataset, 20$\%$ of  records were randomly selected as test data. The rest of the data were used to randomly draw training sets of five given sizes, total 20 random training sets of each size.  All the methods were trained on each training set and tested on the same test set.  The experiment was repeated  10  times starting with selection of the test set. So, every method was applied 200 times on the training sets of the same size. For each method, for each sample size,  average CI on the test data was calculated over all 200 models. 

The Figures 1, 2 demonstrate trends in methods performance with increasing number of instances in the training set. As expected,   average performance of each method improves with the number of training cases. In both cases, Smooth Rank has higher accuracy than two other methods when training sets are small. For the Colon dataset, as the number of training instances grows, CoxPH regression and CoxPath methods surpass in accuracy Smooth Rank.

\begin{figure*}[!tpb] %figure1
\caption{PBC dataset}\label{fig:01}
\centerline{\includegraphics{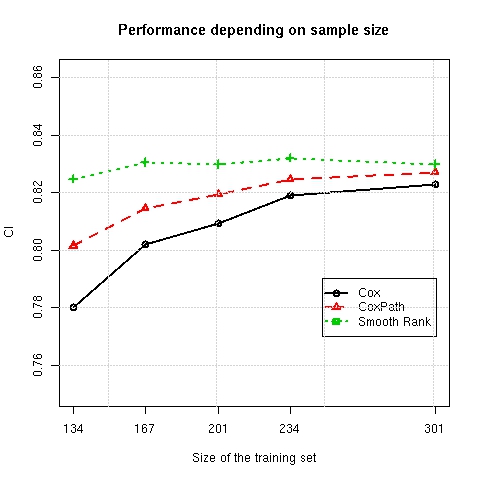}}

\end{figure*}

\begin{figure*}[!tpb] %figure2
\caption{Colon dataset}\label{fig:02}
\centerline{\includegraphics{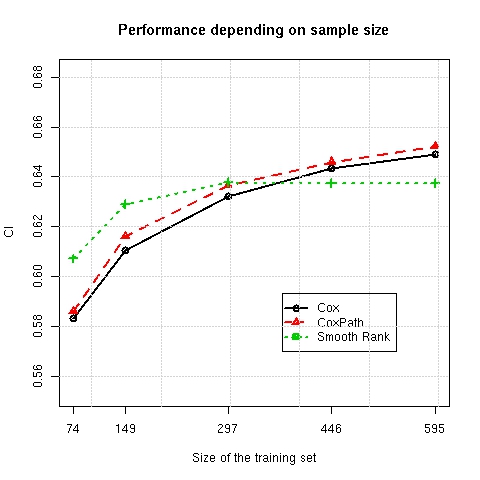}}

\end{figure*}

Overall, the experiment confirms advantage of the Smooth Rank on the scarce data. 

In particularly, on the PBC data,  Smooth Rank achieved the same quality of prediction, as two other methods, with about half of the training instances.  This is an important quality for longitudinal studies, where each individual requires months or years of observations,  making it difficult to increase the number of instances for learning. 

The experiment  confirms our hypothesis that observed higher accuracy of Cox PH regression and CoxPath  on the Colon dataset (see Table 1) can be explained by the larger than usual size of the training sample,  not some specifics of the data. If  only portion of the instances was available, Smooth Rank would be the best method to model the data.

\subsection{Experiments with increasing number of features}

It is very common in survival analysis with medical applications  that, out of all available information,  features for modeling are selected because they are known to have some bearing on the risk of failure. The artificial datasets for this experiment were designed to model this type of data.  

First,  $risk$ was generated as logarithm from a normally distributed random variable. 
With the assumption that times of events $t$ depend on the risk and some unknown factors, the variable $t$  was generated by the formula $t = risk + q$, where $q$ is a random variable with uniform distribution within interval  $ [ min(risk)/2, max(risk)/2]$. 
 
Half of the records were randomly assigned status ``censored''. For non censored observations (events), $target = t.$ For the censored observations, target times indicate the end of observation, which happened before the event (failure) occurred. The target times of the censored observations were calculated by formula $target = t \cdot z$, where $z$ is a random variable uniformly distributed within interval $[0.2, 0.8]$. 

Every feature $f$ was  generated independently the same way as the times of events $t$: $f = risk + q$.

In reality, the features do not always depend on the risk linearly. However,  having more complex features could affect methods differently and make the effects of the dimensionality on the performance less clear.   

All experiments were conducted on the samples with 400 records, which were split on equally sized train and test sets.

For each $M$ multiple of 5 from the interval  [5 , 75],  training and test samples of  with $M$ features we generated  20 times. For each method,  average performance on the test was calculated for each dimension of the data. 

The Fig. 3. shows the results of the experiments.  All the methods have similar performance with the smallest number of features $M = 5$,  which is equal  1/20 of the number of uncensored observations in the training set.  Cox PH regression does not benefit from adding more features.   For CoxPath, top performance was achieved with $M=15$. Both CoxPath and Cox PH regression have tendency of decreasing accuracy with $M > 15$.

\begin{figure*}[!tpb] %figure3
\caption{Atrificial data}\label{fig:03}
\centerline{\includegraphics{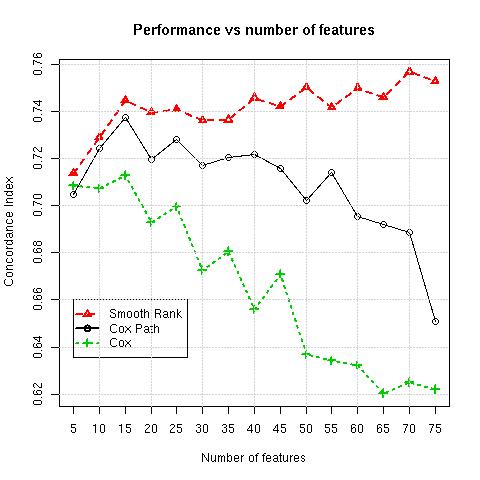}}

\end{figure*}

Smooth Rank is the only method which does not show any sign of overfitting in this experiment. It achieves the best accuracy and the largest advantage over other methods on datasets with highest studied dimensionality $M \ge 70$.   

The difference in accuracy between CoxPath and Smooth Rank for the most scarce datasets is about $0.10$, which is similar to the results we observed in experiments on real data (see Table 1). The consistency between the simulation and  application of the methods on real data may serve to justify the simulation.

\section{Conclusions}

In survival analysis studies with medical applications, it is much easier and cheaper to add features than observations. Unknown factors affect measurements and the outcome to a large degree.  Small, noisy, high-dimensional data put  stringent demands on the robustness of the learning methods. 

We proposed two innovations to address this challenge: (1) reduction of the survival analysis to bipartite ranking; (2) a new robust algorithm for bipartite ranking, Smooth Rank. The method does not use multidimensional optimization to avoid the ``curse of dimensionality''; it uses smoothing techniques while building marginal predictors.  

Advantages of Smooth Rank were proved experimentally, in comparison against the most popular methods for survival analysis (CoxPath and Cox PH regression) in three types of tests. First, the three methods were applied on the 10 real life survival analysis datasets. Then, to systematically study effects of data scarcity on the methods  performance, we conducted two computational experiments:  with real data, where some instances were randomly removed to produce series of training samples of different sizes,   and  with artificial data, where the number of features was gradually increased. 

All three types of experiments, indeed,  demonstrated that Smooth Rank has sizable advantage in accuracy over other two methods on data with smaller number of observations and/or higher dimensionality. The method does not suffer from overfitting where two other methods do. This can make the method a valuable tool in survival analysis studies. 

Smooth Rank is a general bipartite ranking method. Even though its creation was motivated by risk modeling, it may be useful in other applications where robustness of learning is critical. Comparison of the method with  other (bipartite) ranking algorithms on other applications may be a subject of another study.

%\section*{References}
% The bibtex filename
\bibliography{survival}

\end{document}